\documentclass[letterpaper]{article} 
\usepackage{aaai2026}  
\usepackage{times}  
\usepackage{helvet}  
\usepackage{courier}  
\usepackage[hyphens]{url}  
\usepackage{graphicx} 
\usepackage{natbib}  
\usepackage{caption} 
\usepackage{algorithm}
\usepackage{algorithmic}

\usepackage{newfloat}
\usepackage{listings}
\DeclareCaptionStyle{ruled}{labelfont=normalfont,labelsep=colon,strut=off} 
\floatstyle{ruled}
\newfloat{listing}{tb}{lst}{}
\floatname{listing}{Listing}
\usepackage{amsmath}
\usepackage{amsfonts}
\usepackage{multirow}
\usepackage{booktabs}
\usepackage{makecell}
\newcommand\up[1]{{$\uparrow{#1}$}}
\newcommand\down[1]{{$\downarrow{#1}$}}
\nocopyright

\title{BEVDilation: LiDAR-Centric Multi-Modal Fusion for 3D Object Detection}
\author{
    Guowen Zhang\textsuperscript{\rm 1},
    Chenhang He\textsuperscript{\rm 1},
    Liyi Chen\textsuperscript{\rm 1},
    Lei Zhang\textsuperscript{\rm 1}\thanks{Corresponding author}
}
\affiliations{
    \textsuperscript{\rm 1} The Hong Kong Polytechnic University\\
    \{guowen.zhang, liyi0308.chen\}@connect.polyu.hk, chenhang.he@polyu.edu.hk, cslzhang@comp.polyu.edu.hk
}

\usepackage{bibentry}

\begin{document}

\maketitle

\begin{abstract}

Integrating LiDAR and camera information in the bird’s eye view (BEV) representation has demonstrated its effectiveness in 3D object detection.
However, because of the fundamental disparity in geometric accuracy between these sensors, indiscriminate fusion in previous methods often leads to degraded performance.
In this paper, we propose BEVDilation, a novel LiDAR-centric framework that prioritizes LiDAR information in the fusion.
By formulating image BEV features as implicit guidance rather than naive concatenation, our strategy effectively alleviates the spatial misalignment caused by image depth estimation errors.
Furthermore, the image guidance can effectively help the LiDAR-centric paradigm to address the sparsity and semantic limitations of point clouds.
Specifically, we propose a Sparse Voxel Dilation Block that mitigates the inherent point sparsity by densifying foreground voxels through image priors.
Moreover, we introduce a Semantic-Guided BEV Dilation Block to enhance the LiDAR feature diffusion processing with image semantic guidance and long-range context capture.
On the challenging nuScenes benchmark, BEVDilation achieves better performance than state-of-the-art methods while maintaining competitive computational efficiency.
Importantly, our LiDAR-centric strategy demonstrates greater robustness to depth noise compared to naive fusion.
    
\end{abstract}

\begin{links}
    \link{Code}{https://github.com/gwenzhang/BEVDilation}
\end{links}

\section{Introduction}
\label{sec:intro}

3D object detection plays an important role in applications of autonomous driving~\cite{e2ead_survey}, virtual reality~\cite{VR_3Ddetection}, and editing~\cite{liyi2025vip3de}. 
To ensure precise and robust perception, LiDAR-camera fusion~\cite{huang2024DAL,liu2023bevfusion-mit,liang2022bevfusion-pku,yang2022deepinteraction,wang2023unitr,sparsefusion} is employed to integrate complementary information.
In specific, LiDAR point clouds can offer accurate location and geometry information, while RGB images provide rich semantic and context information.
However, significant view discrepancies and inherent modality limitations make robust and effective 3D multi-modal detection particularly challenging.
To tackle this challenge, recent works focus on establishing cross-model correspondences to enable unified representation learning.
Input-based~\cite{pointpainting,wang2021pointaugmenting,yin2021multimodal} and query-based methods~\cite{yang2022deepinteraction,sparsefusion,yan2023CMT,transfusion} enhance LiDAR point clouds or object proposals by integrating semantic information from RGB images. BEV-based methods~\cite{liu2023bevfusion-mit,liang2022bevfusion-pku,yin2024isfusion,jiao2023msmdfusion} align and fuse multi-modal features in a unified BEV representation.

\begin{figure}[t]
\centering
\includegraphics[width=0.45\textwidth]{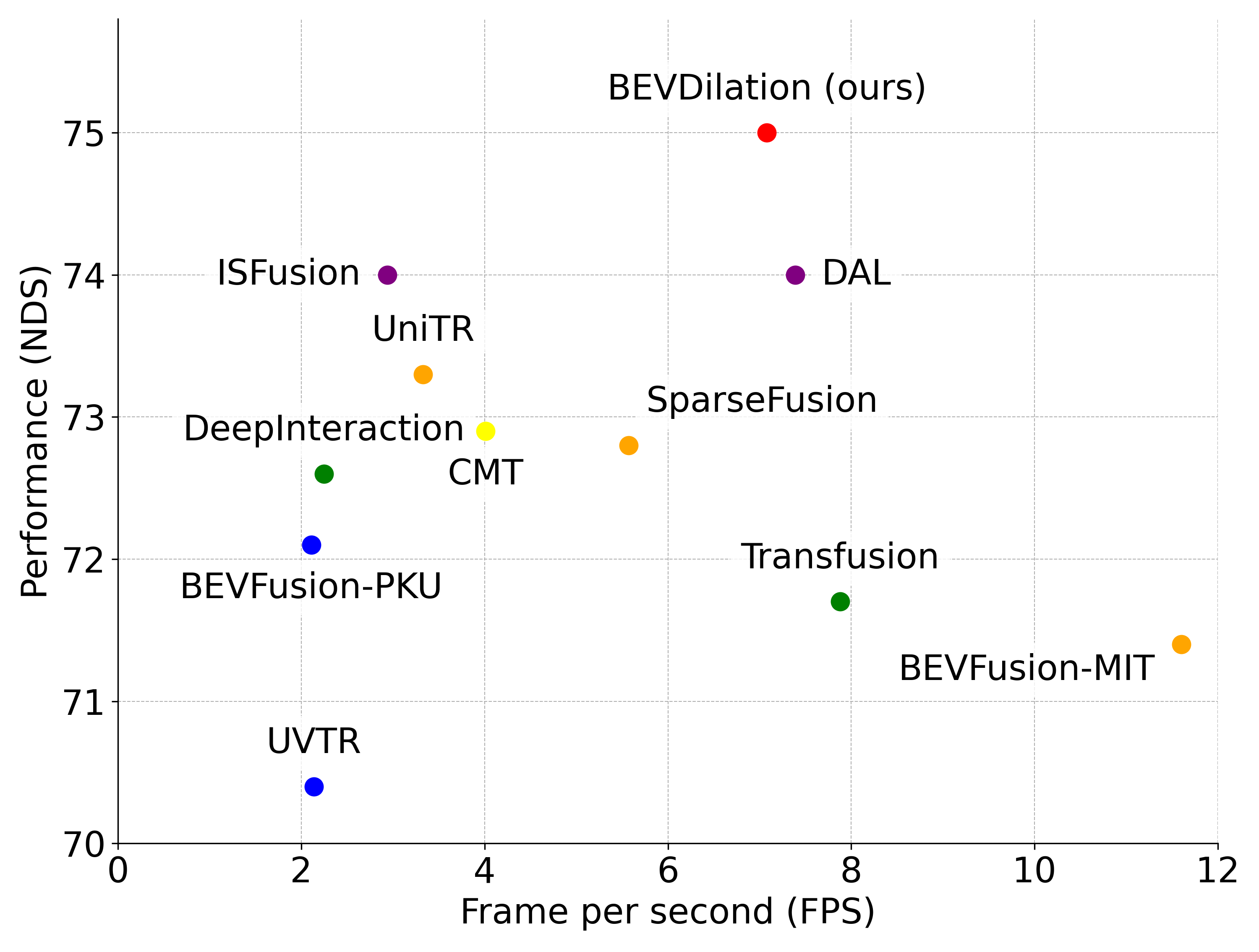}
\caption{3D object detection performance (NDS) vs speed (FPS) on nuScenes validation set. 
}
\label{Fig::speed_performance}
\end{figure}

\begin{figure}[t]
\centering
\includegraphics[width=0.4\textwidth]{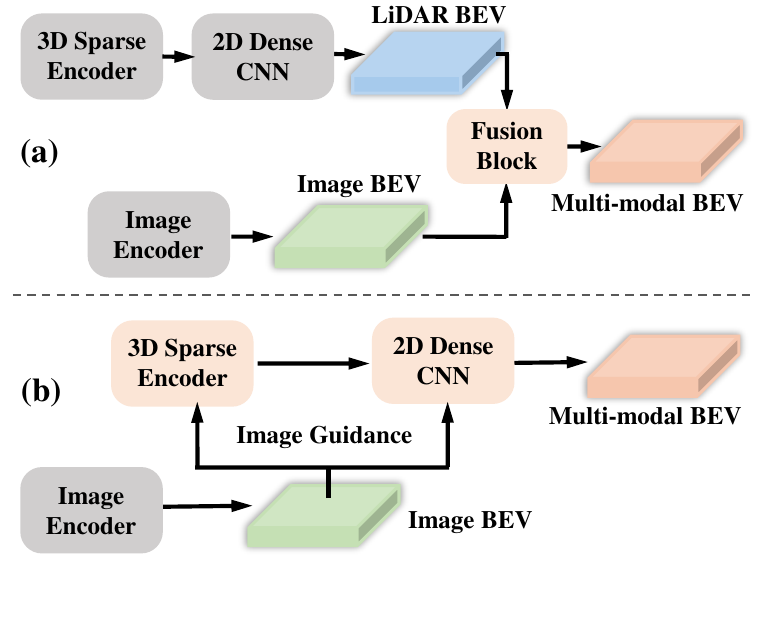}
\caption{Comparison of (a) indiscriminate fusion and (b) our LiDAR-centric strategy.}
\label{Fig::Insight}
\end{figure}

However, the fundamental disparity in geometric and localization accuracy between LiDAR and cameras implies that indiscriminate fusion can lead to degraded performance~\cite{huang2024DAL}.
Specifically, cameras lack dense depth priors, and depth estimation methods (e.g., LSS~\cite{philion2020LSS}) are prone to errors and overfitting~\cite{li2023bevdepth,huang2024DAL}.
In contrast, point clouds provide accurate localization information.
The superior performance of LiDAR-based detectors~\cite{zhang2025voxelmamba,liu2025lion} over image-based detectors~\cite{li2024bevformer,liu2022petr} further demonstrates this inherent disparity.
Thus, as shown in Fig~\ref{Fig::Insight} (a), naively aligning point cloud features with those derived from estimated depth constitutes a suboptimal fusion paradigm.
Given the localization accuracy gap between point clouds and images, LiDAR information should be prioritized.
However, this prioritization in multi-modal fusion is non-trivial.
First, prioritizing LiDAR also highlights its own limitations.
Point clouds are highly sparse and unevenly distributed, leading to incomplete geometric information and constrained representation capacity.
Besides, in the LiDAR modality, object centers are usually empty, which weakens the representation power of point clouds. 
To address center feature missing~\cite{FSD}, previous methods~\cite{hednet,Voxelnet,Second} often use feature diffusion~\cite{FSD} (\textit{e.g.,} stacked 2D CNN), which struggles to capture context information and distinguish noise due to insufficient semantic information.
Second, when integrating image information, it is challenging to design effective fusion mechanisms to preserve the primacy of LiDAR data, thereby reducing the noise from image depth estimation.

To overcome the aforementioned limitation, in this paper we propose a novel LiDAR-centric fusion paradigm, termed BEVDilation, for multi-modal 3D object detection. 
As illustrated in Fig~\ref{Fig::Insight} (b), by formulating the image features as guidance, the fused multi-modal features can primarily derive their geometric and localization information from LiDAR point clouds. 
Specifically, image features provide semantic and contextual cues to drive sparse-to-dense feature diffusion in the LiDAR detector, rather than being naively aligned with point cloud features.
This strategy reduces the impact of image depth estimation noise and improves fusion robustness.
To address the inherent limitations of point clouds, we further propose two modules with it.
Inspired by voxel generation~\cite{liu2025lion,swformer,zhang2024safdnet} in LiDAR-based detectors, which do not rely on strict geometric information, we introduce the {\textbf{S}}parse {\textbf{V}}oxel {\textbf{D}}ilation {\textbf{B}}lock (SVDB) to densify foreground features and enhance features representation in highly sparse point clouds.
It employs complementary multi-modal features to identify the foreground regions in BEV space and fills LiDAR’s empty foreground areas with new learnable voxels.
This incorporates the image instance information into LiDAR and improves the completeness of point cloud features.
To capture long-range context dependencies and refine point features, we propose the {\textbf{S}}emantic-Guided {\textbf{B}}EV {\textbf{D}}ilation {\textbf{B}}lock (SBDB) for LiDAR feature diffusion.
In specific, following deformable convolution~\cite{dai2017DCN,zhu2019DCNv2}, we add offsets to the regular convolution sampling position to improve feature diffusion efficiency and adapt to geometric variation.
Furthermore, SBDB uses modulation scalars to filter out noise features based on image semantic information.
All deformations are conditioned on multi-modal BEV features but operate solely on LiDAR features.
In this way, better LiDAR representations can be achieved under our LiDAR-centric design.

In summary, the major contributions of our work are:
\begin{itemize}
    \item We propose BEVDilation, a LiDAR-centric framework for multi-modal 3D detection. BEVDilation formulates image features as guidance while prioritizing LiDAR information, achieving effective and robust fusion.
    \item To address the limitations of point clouds in the LiDAR-centric paradigm, we propose the Sparse Voxel Dilation Block (SVDB) and Semantic-Guided BEV Dilation Block (SBDB) to reduce sparsity and augment semantic information in BEVDilation.
    \item Extensive experiments on the nuScenes~\cite{Nuscenes} dataset show that our BEVDilation yields a new state-of-the-art for multi-modal 3D object detection while maintaining competitive computational efficiency.
\end{itemize}

\section{Related Work}
\label{sec:related_work}

\textbf{LiDAR-based 3D Object Detection.}
LiDAR sensors are the most widely adopted sensor in autonomous driving due to the precise geometric information provided by point clouds. Current LiDAR-based 3D object detection methods rely on two major representations: point clouds and voxels. Inspired by PointNet~\cite{Pointnet}, point-based methods~\cite{Pointrcnn,Votenet} directly extract geometric features from local regions of raw point clouds. 
However, these methods suffer from low inference efficiency and insufficient context modeling due to the irregular and sparse nature of point clouds. 
Voxel-based methods~\cite{scatterformer,zhang2025voxelmamba,liu2025lion,GGA,VoxelTransformer,zhao2024simdistill} transform raw point clouds into regular grids and well balance performance and efficiency. Voxel-based methods currently dominate the field of LiDAR-based 3D object detection. 
They often adopt a sparse backbone~\cite{zhang2025voxelmamba,Second} to extract features from sparse voxels in 3D space. These features are then converted to dense feature maps in BEV and processed by a 2D dense backbone~\cite{Second,hednet} to enlarge the receptive field and refine features.
However, LiDAR-based detectors relying solely on sparse and irregular point clouds exhibit degraded performance in challenging scenarios, such as long-range and geometrically similar object (\textit{e.g.,} pedestrians and vertical poles) detection.
In this paper, we integrate additional image information to address the limitation of point cloud data by densifying sparse features and incorporating semantic guidance.

\noindent \textbf{Multi-Modality 3D Object Detection.}
To achieve accurate and robust perceptions, existing approaches~\cite{sparsefusion,wang2023unitr,huang2024DAL,yin2024isfusion,yan2023CMT,liu2023bevfusion-mit} explore capturing complementary information from LiDAR and cameras. Despite the substantial performance gap between camera-based~\cite{liu2022petr,liu2023petrv2,li2024bevformer,li2023bevdepth,man2025locateanything3d} and LiDAR-based detectors~\cite{zhang2025voxelmamba,liu2025lion}, the rich semantic information in images can further enhance LiDAR representations. 
Current multi-modal 3D detection approaches can be basically classified into three categories: input-based~\cite{pointpainting,wang2021pointaugmenting,yin2021multimodal}, BEV-based~\cite{liu2023bevfusion-mit,liang2022bevfusion-pku,yin2024isfusion,huang2024DAL} and query-based~\cite{yan2023CMT,sparsefusion,yang2022deepinteraction}. Pioneering input fusion methods~\cite{pointpainting,yin2021multimodal} decorate or densify LiDAR points with the semantic information from images, which is more sensitive to calibration errors.
BEV-based fusion~\cite{liu2023bevfusion-mit,liang2022bevfusion-pku,huang2024DAL} methods project LiDAR and image features into a unified BEV representation with estimated depth. 
Nevertheless, the uncertain depth of images introduces spatial misalignment and reduces the robustness of BEVFusion.
Query-based methods~\cite{sparsefusion,yan2023CMT,cai2023objectfusion} extract sparse instance-level representations and refine them with an attention operation.
However, this strategy leaves several fundamental limitations of individual modalities unaddressed.
In contrast, our proposed BEVDilation resolves LiDAR sparsity and semantic deficiency while alleviating spatial misalignment issues.

\section{Method}
\label{sec:method}

In this section, we present BEVDilation, a LiDAR-centric 3D backbone that can be applied for multi-modal 3D detection. 
First, we introduce the overall architecture of BEVDilation.
Then, we describe in detail the fundamental components of BEVDilation, including the Sparse Voxel Dilation Block and Semantic-Guided BEV Dilation Block.

\begin{figure*}[t]
     \centering
    \includegraphics[width=1.0\textwidth]{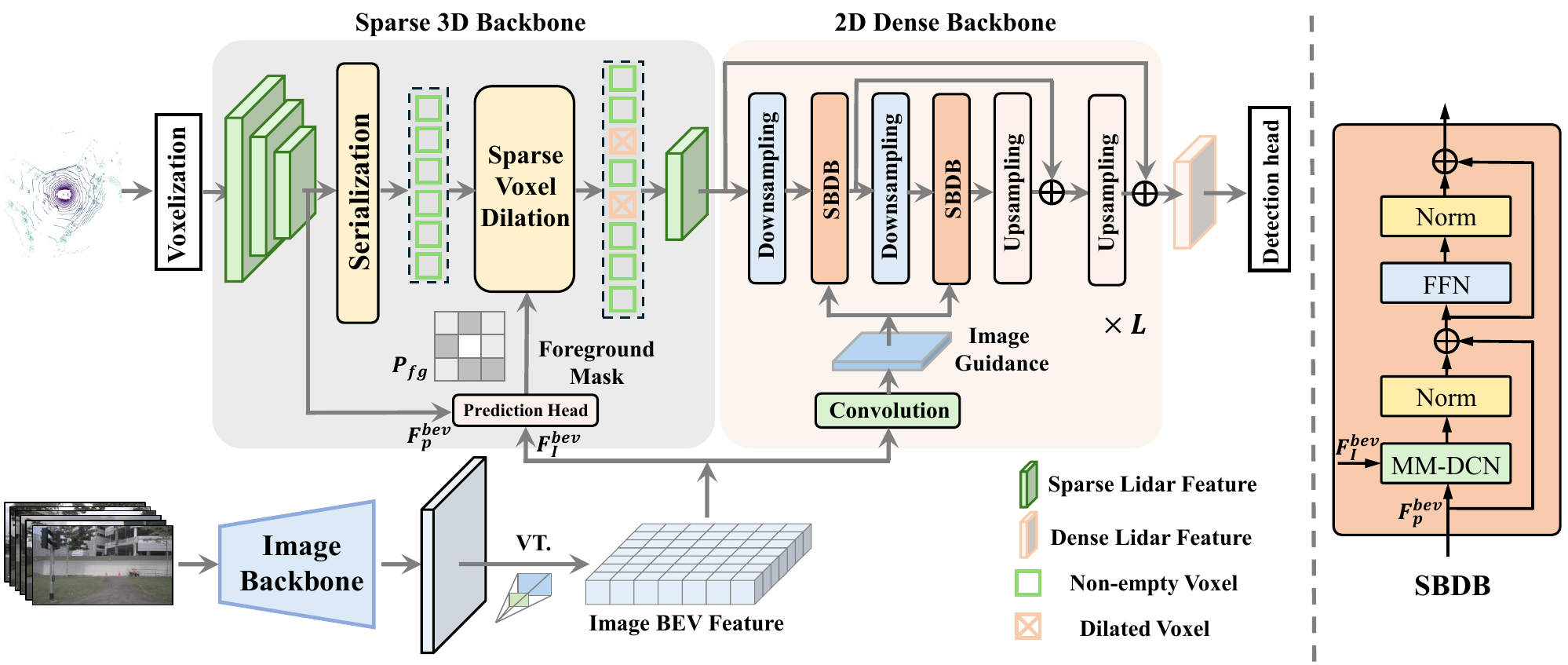}
     \caption{ The overall architecture of our proposed \textbf{BEVDilation}. Given the point clouds and multi-view images, we take two individual backbones to extract multi-modal BEV features. For the LiDAR branch, we enhance the LiDAR BEV features with our proposed Sparse Voxel Dilation Block and Semantic-Guided BEV Dilation Block.
     }
    \label{fig:main}
\end{figure*}

\subsection{Overall Architecture}
An overview of our proposed BEVDilation is shown in Fig~\ref{fig:main}. 
As in previous works~\cite{transfusion,sparsefusion,yin2024isfusion}, BEVDilation takes LiDAR point clouds and corresponding multi-view images as input, which are fed into two individual backbones to extract multi-modal BEV features. Unlike previous multi-modal methods that indiscriminately fuse modality-specific features~\cite{liu2023bevfusion-mit,liang2022bevfusion-pku,sparsefusion,yan2023CMT}, BEVDilation introduces a LiDAR-centric framework. It uses image features as guidance to mitigate the inherent limitations of point clouds, including sparsity, irregular distributions, and semantic ambiguities. Specifically, a Sparse Voxel Dilation Block working on the sparse voxel features is proposed, which increases voxel density in foreground instances and completes features in LiDAR-occluded areas.
Image and LiDAR features are fused to generate a BEV foreground mask, which subsequently determines the regions for voxel dilation.
To mitigate the semantic ambiguity and improve geometric accuracy, BEVDilation introduces a Semantic-Guided BEV Dilation Block that integrates multi-modal guidance in feature diffusion, transforming sparse LiDAR BEV features into geometrically consistent and semantically enriched dense LiDAR representations.

\begin{figure}
    \centering
	\includegraphics[width=0.85\columnwidth]{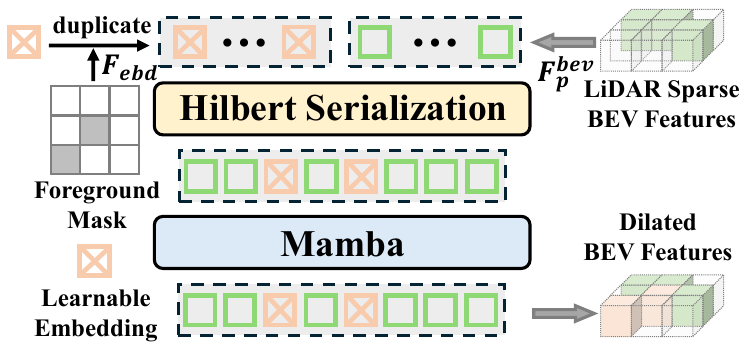}
    \caption{
	An illustration of SVDB. The newly padded and original voxels are merged with global receptive fields.}
	\label{fig:SVDB}
\end{figure}

\subsection{Sparse Voxel Dilation Block}
\label{sec:MVDB}
Given the challenges of high sparsity of point clouds and information degradation in LiDAR modality, we propose a sparse voxel dilation strategy that leverages the autoregressive capacity and global receptive field of Mamba~\cite{liu2025lion} to address these issues. 

\noindent \textbf{Image Encoder.} For each image $I$, a shared image backbone is used to extract multi-scale feature maps that are subsequently fused into a single-scale representation $\mathbf{F}_{I}\in \mathbb{R}^{C_{i}\times H\times W}$ by FPN~\cite{lin2017FPN}, where $H$ and $W$ denote the image resolution and $C_{i}$ represents the number of feature channels. Then, the multi-view image features are transformed from the perspective view to BEV representation $\mathbf{F}_{I}^{bev}\in \mathbb{R}^{C_{ib}\times 1\times Y \times X}$ using the well-established view transformation (VT.) algorithm Lift-Splat-Shot (LSS)~\cite{philion2020LSS}, where $X$ and $Y$ denote the size of the BEV feature map and $C_{ib}$ is the number of image BEV feature channels.

\noindent \textbf{Voxel Sparse Encoder.} For a fair comparison, we use the popular sparse backbone VoxelNet~\cite{Voxelnet} to extract sparse voxel representations. The irregular point clouds $P$ are first transformed into sparse voxels by a voxel feature encoding. Then, we extract sparse voxel features $\mathbf{F}_{P}\in \mathbb{R}^{N\times C_{p}} $ along with their spatial coordinates $\mathcal{V}\in \mathbb{R}^{N\times 3}$ using a series of sparse convolutional blocks. $N$ indicates the number of voxels and $C_{p}$ represents the number of voxel feature channels.

\noindent \textbf{Sparse Voxel Dilation Block.} As shown in Fig~\ref{fig:main}, to enhance the accuracy of dilation regions, we use multi-modal information to predict the foreground mask.
The prediction head takes the image BEV features $\mathbf{F}_{I}^{bev}$ and LiDAR sparse BEV features $\mathbf{F}_{P}^{bev}$ as inputs. 
Specifically, the foreground mask predictor is formulated as follows:
\begin{equation}
\mathbf{P}_{fg} = \sigma(f_{conv}((\mathbf{F}_{I}^{bev};\mathbf{F}_{P}^{bev}))), \\
\end{equation}
where $f_{conv}$ denotes two-layer convolutions, $\sigma$ represents the sigmoid activation, and $(;)$ indicates concat operation. 
The binary foreground mask $\mathcal{M}\in \{0,1\}^{Y\times X}$ is obtained:
\begin{equation}
\mathcal{M}_{i,j} = \mathbb{I}[\mathbf{P}_{fg}(i,j) > \tau],
\end{equation}
where (i, j) denotes the indices within the BEV grid and $\tau$ is the threshold.
Regions corresponding to positive mask values are designated for padding new voxels.
During training, we treat BEV grid cells within object bounding boxes as the ground truth foreground to supervise $\mathcal{M}$.
The dilated voxels can effectively compensate for LiDAR-occluded regions and densify the foreground instances.
Besides, the dilated features in previous methods~\cite{swformer,zhang2024safdnet,liu2025lion} are often assigned using K-NN approaches or zero initialization.
These handcrafted strategies tend to yield sub-optimal results due to their limited adaptability to varying scene contexts.
Therefore, as shown in the Fig~\ref{fig:SVDB}, all newly dilated features are generated as learnable embeddings $\mathbf{F}_{ebd}$.
Inspired by the recent success of Mamba in LiDAR detectors~\cite{zhang2025voxelmamba,liu2025lion}, we utilize its auto-regressive property to refine voxel features with global receptive fields. 
In specific, we merge and sort the newly dilated and original voxels using Hilbert serialization~\cite{hilbertcurve}, and then refine the features with a single group-free Mamba layer~\cite{zhang2025voxelmamba} as following:
\begin{equation}
\mathbf{F}_{P}^{bev} = \textbf{Mamba}(\textbf{HS}([\mathbf{F}_{P}^{bev}; \mathbf{F}_{ebd}])),
\end{equation}
where HS denotes the operation that sorts the voxel sequence using a Hilbert curve~\cite{zhang2025voxelmamba}. Overall, SVDB effectively tackles point cloud sparsity and LiDAR blind spots while preserving LiDAR's geometric primacy.

\subsection{Semantic-Guided BEV Dilation Block}
\label{sec:SFDB}
Though SVDB mitigates foreground sparsity, the BEV feature remains sparse, thereby demanding an efficient feature diffusion~\cite{FSD}. The feature diffusion in existing LiDAR detectors exhibits two major limitations: semantic ambiguity and the rigid geometric constraints in standard convolutions. Therefore, we introduce the Semantic-Guided BEV Dilated Block to build a semantic-guided feature diffusion and enhance LiDAR's ability to model geometric variations through multi-modal deformation.

As shown in Fig.~\ref{fig:main}, the SBDB is designed with a multi-modal deformable convolution (MM-DCN), a feed-forward network (FFN), layer normalization, and residual connections.
To enlarge the receptive field and improve computational efficiency, it operates on downsampled BEV features.
In MM-DCN, the deformations, specifically the offset and modulation scalar, are conditioned on the multi-modal BEV information.
Moreover, to reduce spatial misalignment from uncertain depth estimation, while the deformations (offsets and modulation scalars) in MM-DCN are conditioned on multi-modal BEV features, the deformable convolutions operate solely on LiDAR features.
Specifically, image BEV features are first downsampled using a lightweight convolutional encoder to match the resolution of LiDAR BEV features.
These downsampled image BEV features are shared across different stages.
Subsequently, the SBDB takes these processed image BEV features $\mathbf{F}_{I}^{bev}$ and the corresponding LiDAR BEV features $\mathbf{F}_{P}^{bev}$ as input, defined as:
\begin{equation}
\begin{aligned}
    &\widetilde{\mathbf{F}}^{bev}_{P} = \textbf{LN}(\textbf{MM-DCN}((\mathbf{F}_{P}^{bev}, \mathbf{F}_{I}^{bev}))) + \mathbf{F}_{P}^{bev}, \\
    &\mathbf{F}_{P}^{bev}=\textbf{LN}(\textbf{MLP}(\widetilde{\mathbf{F}}^{bev}_{P})) + \widetilde{\mathbf{F}}^{bev}_{P}.
\end{aligned}
\end{equation}
Here, $\textbf{LN}(\cdot)$ stands for Layer Normalization.
$\textbf{MM-DCN}(\cdot)$ represents the deformable convolutions used in the first residual branch, where deformations (offsets $\Delta p_k$ and modulation scalars $\textbf{m}_{k}$) are conditioned on multi-modal features. The convolution is applied solely to the LiDAR features $\mathbf{F}_{P}^{bev}$.
Formally, the \textbf{MM-DCN} is denoted as:
\begin{equation}
\hat{\mathbf{F}}_{P}^{bev}(p_{0}) = \sum_{k=1}^{M} \textbf{w}_{k}\textbf{m}_{k} \cdot \mathbf{F}_{P}^{bev}\bigl(p_{0} + p_k + \Delta p_k),
\end{equation}
where $\mathbf{F}_{P}^{bev}(p)$ denotes the features at location $p$ from LiDAR BEV features, and $\textbf{w}_{k}$ are the standard convolutional kernel weights for the k-th location. $M$ represents the total number of sampling points.
For clarity, groups in DCN~\cite{wang2023DCNv3} are omitted.
By adapting multi-modal deformable convolutions with the LiDAR-centric strategy, SBDB achieves precise and robust feature diffusion, enhancing the LiDAR representations.

\subsection{Integration with Detection}
\label{sec:Pipeline}
With the proposed SVDB and SBDB, we build BEVDilation, a LiDAR-centric multi-modal fusion paradigm. The architecture of BEVDilation is illustrated in Fig~\ref{fig:main}. The SVDB is placed subsequent to the VoxelNet sparse backbone. Our 2D dense backbone, comprising $L$ stages, directly replaces the dense backbone of the baseline.
For the detection head and loss function, we adopt the same settings as DAL~\cite{huang2024DAL}, adding a focal loss to supervise the binary foreground mask.

\section{Experiments}

\begin{table*}[t]
    \centering
    \setlength{\tabcolsep}{1mm} 
    \begin{tabular}{l|c|cc|cc|cc|c}
    \toprule
        \multirow{2}{*}{Methods} & \multirow{2}{*}{Modality} & LiDAR & Camera & \multicolumn{2}{c|}{\textit{validation set}} & \multicolumn{2}{c}{\textit{test set}} & \multirow{2}{*}{FPS}\\
        & & Backbone & Backbone & NDS & mAP & NDS & mAP &\\
        \midrule
        PETRv2~\cite{liu2023petrv2} & C & - & ResNet-101 & 52.4 & 42.1 & 55.3 & 45.6 & -\\
        BEVFormer~\cite{li2024bevformer} & C & - & ResNet-101 & 51.7 & 41.6 & 53.5 & 44.5 & - \\
        \midrule
        SECOND~\cite{Second} & L & VoxelNet & - & 63.0 & 52.6 & 63.3 & 52.8 & - \\
        TransFusion-L~\cite{transfusion}& L & VoxelNet & - & 70.1 & 65.1 & 70.2 & 65.5 & - \\ 
        FocalFormer3D-L~\cite{focalformer3d} & L & VoxelNet & - & 70.9 & 66.4 & 72.6 & 68.7 & - \\
        \midrule
        MVP~\cite{yin2021multimodal} & L+C & VoxelNet & CenterNet & 70.0 & 66.1 & 70.5 & 66.4 & - \\
        GraphAlign~\cite{song2023graphalign} & L+C & VoxelNet & DeepLabv3 & - & - & 70.6 & 66.5& - \\
        FUTR3D~\cite{chen2023futr3d} & L+C & VoxelNet & ResNet-101 & 68.3 & 64.5 & - & - & -\\
        UVTR~\cite{UVTR} & L+C & VoxelNet & ResNet-101 & 70.2 & 65.4 & 71.1 & 67.1 & 2.14\\
        TransFusion~\cite{transfusion} & L+C & VoxelNet & ResNet-50 & 71.3 & 67.5 & 71.6 & 68.9 & 7.88\\
        AutoAlignV2~\cite{chen2022autoalignv2} & L+C & VoxelNet & CSPNet & 71.2 & 67.1 & 72.4 & 68.4 & -\\
        BEVFusion~\cite{liang2022bevfusion-pku} & L+C & VoxelNet & Dual-Swin-T & 72.1 & 69.6 & 73.3 & 71.3 & 2.11\\
        BEVFusion~\cite{liu2023bevfusion-mit} & L+C & VoxelNet & Swin-T & 71.4 & 68.5 & 72.9 & 70.2 & \textbf{11.60}\\
        DeepInteraction~\cite{yang2022deepinteraction} & L+C & VoxelNet & ResNet-50 & 72.6 & 69.9 & 73.4 & 70.8 & 2.25\\
        CMT~\cite{yan2023CMT} & L+C & VoxelNet & VoV-99 & 71.9 & 69.4 & 73.0 & 70.4 & 4.01\\
        FocalFormer3D~\cite{focalformer3d} & L+C & VoxelNet & ResNet-50 & 73.1 & 70.5 & 73.9 & 71.6 & - \\
        MSMDFusion~\cite{jiao2023msmdfusion} & L+C & VoxelNet & ResNet-50 & - & - & 74.0 & 71.5 & 2.57 \\
        SparseFusion~\cite{xie2023sparsefusion} & L+C & VoxelNet & ResNet-50 & 72.8 & 70.4 & 73.8 & 72.0 & 5.57\\
        UniTR~\cite{wang2023unitr} & L+C & DSVT & DSVT & 73.3 & 70.5 & 74.5 & 70.9 & 3.33$^{\dag}$\\
        ObjectFusion~\cite{cai2023objectfusion} & L+C & VoxelNet & Swin-T & 72.3 & 69.8 & 73.3 & 71.0 & - \\
        GraphBEV~\cite{song2024graphbev} & L+C & VoxelNet & Swin-T & 72.9 & 70.1 & 73.6 & 71.7 & 3.12\\
        DAL~\cite{huang2024DAL} & L+C & VoxelNet & ResNet-50 & 74.0 & 71.5 & 74.8 & 72.0 & 7.39\\
        \midrule
        \textbf{BEVDilation} (ours) & L+C & VoxelNet & ResNet-50 & \textbf{75.0} & \textbf{73.0} & \textbf{75.4} & \textbf{73.1} & 7.08\\
        \bottomrule
    \end{tabular}
    \caption{Comparison with existing methods on nuScenes \textbf{validation} and \textbf{test} set. `L' denotes the LiDAR and 'C' indicates the camera. Symbol ${\dag}$ means UniTR without cache acceleration. The inference speeds are evaluated on an NVIDIA A6000 GPU.}  
    \label{tab:SOTA}
\end{table*}

\subsection{Datasets and Evaluation Metrics}
Following previous works~\cite{sparsefusion,yan2023CMT,transfusion}, we evaluate our method on the nuScenes dataset~\cite{Nuscenes}. This dataset comprises 750 training scenes, 150 validation scenes, and 150 testing scenes. It provides point clouds from a 32-beam LiDAR and 6 images from multi-view cameras. For 3D object detection, nuScenes employs the mean Average Precision (mAP), nuScenes detection scores (NDS), mean Average Translation Error (mATE), mean Average Scale Error (mASE), mean Average Orientation Error (mAOE), mean Average Velocity Error (mAVE) and mean Average Attribute Error (mAEE) to measure model performance.

\subsection{Implementation Detail}
Our method is implemented based on the open-source framework MMDetection3d~\cite{mmdet3d}. For the camera branch, we use ResNet50~\cite{ResNet} as the image backbone and initialized it with weights pre-trained on nuImage~\cite{Nuscenes}, following current approaches~\cite{yang2022deepinteraction,yin2024isfusion,sparsefusion}. The input image resolution is 800$\times$448.
For the LiDAR branch, we follow DAL~\cite{huang2024DAL} and use the voxel size $(0.075m, 0.075m, 0.2m)$, with the point cloud range defined as $[-54m, 54m]$ along the XY-axes and $[-5m, 3m]$ for the Z-axis. The initial BEV feature map is of size $180\times 180$. 
We adopt the sparse 3D backbone from VoxelNet~\cite{Voxelnet}, consistent with previous works~\cite{sparsefusion,huang2024DAL,yin2024isfusion}. 
For the 2D dense CNN, we stack eight SBDB blocks distributed over four stages.
The $\tau$ in SVDB is set to 0.4.
The number of groups in the multi-modal deformable convolutions~\cite{wang2023DCNv3} is set to 16.

\noindent \textbf{Training and Inference.} Following previous training schemes~\cite{yin2024isfusion,huang2024DAL}, BEVDilation is trained in an end-to-end manner with one stage. We optimize the model using AdamW optimizer with weight decay $0.01$, one-cycle learning rate policy, max learning rate 0.0001, and batch size 24 for 10 epochs. The class-balanced sampling strategy from CBGS~\cite{zhu2019CBGS} and multi-modal data augmentation from DAL~\cite{huang2024DAL} are adopted during training. 
We use the CenterPoint~\cite{Centerpoint} detection head.
All the models are trained on 8 RTX A6000 GPUs.
During evaluation, image features are only used for searching proposal candidates, while regression relies entirely on LiDAR features as in~\cite{huang2024DAL}.

\subsection{Comparison with State-of-the-art Methods}

As shown in Table~\ref{tab:SOTA}, we compare BEVDilation with state-of-the-art (SOTA) methods on both the validation and test set of the nuScenes dataset without any test-time augmentations or model ensembles. Our proposed BEVDilation achieves impressive results with 75.0 NDS and 73.0 mAP on the validation set, which is at least +1.0 and +1.5 higher than the SOTA method, including those with more powerful backbones. 
Our method also exhibits the best mAP and NDS on the test set. 
Additionally, as shown in Fig~\ref{Fig::speed_performance}, BEVDilation strikes an effective balance between speed and accuracy. 
It outperforms DAL +1.0 NDS in detection accuracy on the nuScenes validation while achieving comparable speed. 
Some methods, like BEVFusion, are faster than BEVDilation; however, their accuracy is substantially lower.
A more detailed performance analysis, as well as additional experimental results, is provided in the \textbf{supplementary material}.

Following common model scaling practices~\cite{huang2024DAL,sparsefusion}, we enhance BEVDilation with Swin-Tiny~\cite{swin_transformer} and a higher input resolution.
Table~\ref{tab:scale} presents a comparison between the scaled BEVDilation and other scaled SOTA methods.
Despite Mambafusion employing more advanced LiDAR (LION-Mamba) and image (VMamba) backbones, BEVDilation nonetheless demonstrates significant advantages, surpassing it by +0.3 NDS and +0.7 mAP while achieving faster inference speeds. 
BEVDilation also maintains superior computational efficiency compared to most existing approaches.


  \begin{table}[t]
     \centering
      \setlength{\tabcolsep}{1mm} 
     \begin{tabular}{lccccc}
        \toprule
        Methods & \makecell{Image\\Backbone} & \makecell{Input\\Resolution} & NDS & mAP & FPS\\
        \midrule
        CMT & VoVNet-99 & 1600$\times$640 & {72.9} & 70.3 & 4.01\\
        SparseFusion & Swin-T & 800$\times$448 & 73.1 & {71.0} & 5.22 \\
        DAL & ResNet-50 & 1056$\times$384 & 74.0 & 71.5 & \textbf{7.39} \\
        IS-Fusion & Swin-T & 1056$\times$384 &  74.0 & 72.8 & 2.94 \\
        MambaFusion & VMamba & 704$\times$256 & 75.0 & 72.7 & 3.84 \\
        \midrule
        BEVDilation & Swin-T & 1056$\times$384 & \textbf{75.3} & \textbf{73.4} & 4.13\\
        \bottomrule
    \end{tabular}
    \caption{Results on nuScenes \textbf{validation} set with stronger image backbone and higher input resolution. }
    \label{tab:scale}
  \end{table}

  \begin{table}[t]
     \centering
      \setlength{\tabcolsep}{1mm}
        {
        \begin{tabular}{ccc|lll}
        \toprule
         Baseline-LC & SVDB & SBDB  & mAP & NDS & FPS \\ 
        \midrule
         \checkmark           &     &   &  70.6  & 73.3  & 9.12      \\  
        \midrule
         \checkmark          &  \checkmark   &   & 71.8  & 74.0 & 8.62         \\
         \checkmark          &     &  \checkmark & 72.5     &  {74.8} & 7.51     \\ 
        \checkmark           &  \checkmark   &  \checkmark &\textbf{73.0}  &\textbf{75.0} & 7.08  \\
        \bottomrule
        \end{tabular}
        }
    \caption{Ablations on the nuScenes validation split. Baseline-LC denotes our multi-modal baseline method.}
    \label{tab:eachcomp}
  \end{table}

\begin{figure*}[t]
     \centering
    \includegraphics[width=0.83\textwidth]{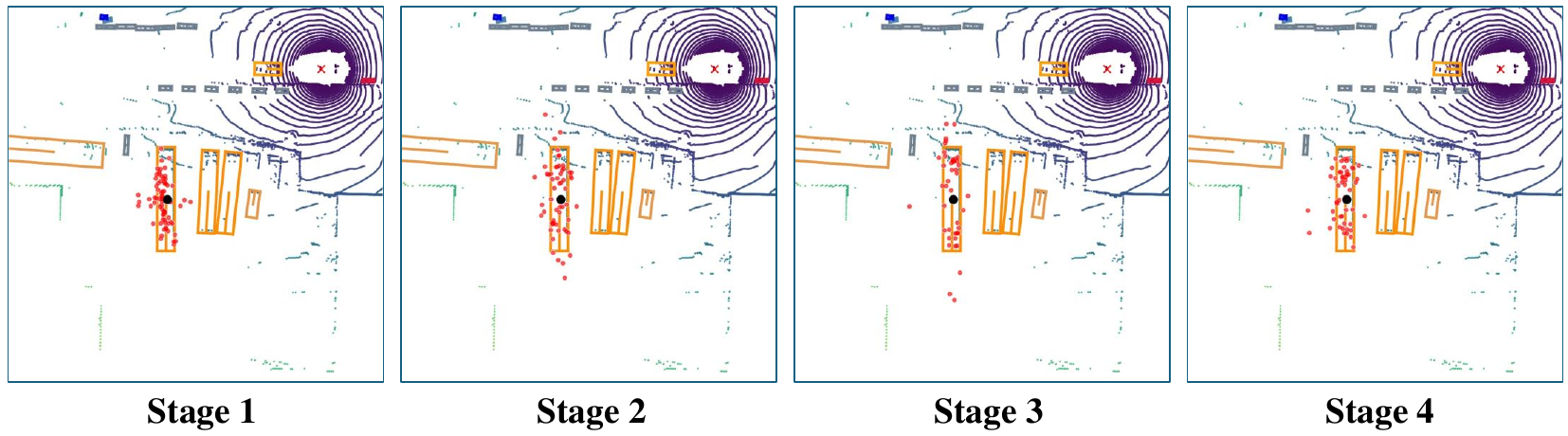}
     \caption{ Visualization of sampling locations of \textbf{SBDB} at different stages. The black dot indicates the object center, and these red dots denote the sampling locations of this center query point in SBDB. 
     }
    \label{fig:deformation}
\end{figure*}

\begin{table*}[t]
    \centering
     \setlength{\tabcolsep}{1mm}
    {
        \begin{tabular}{l|cccccccc}
        \toprule
        \multirow{2}{*}{Methods} & \multicolumn{3}{c}{BEVDilation} & \multicolumn{3}{c}{BEVFusion} \\
        & mAP$\uparrow$ & NDS$\uparrow$ & mATE$\downarrow$ & mAP$\uparrow$ & NDS$\uparrow$ & mATE$\downarrow$\\ 
        \midrule
        Baseline & 73.0 & 75.0 & 26.9  & 68.5 & 71.4 & 28.7 \\
        One-hot Noise & 70.3 \down{\textit{\footnotesize 2.7}} & 73.6 \down{\textit{\footnotesize 1.4}} & 27.2 \up{\textit{\footnotesize 0.3}} & 63.1 \down{\textit{\footnotesize 5.4}} &68.4 \down{\textit{\footnotesize 3.0}} & 29.4 \up{\textit{\footnotesize 0.7}} \\
        Random Noise & 70.0 \down{\textit{\footnotesize 3.0}} & 73.4 \down{\textit{\footnotesize 1.6}} & 27.3 \up{\textit{\footnotesize 0.4}} & 62.3 \down{\textit{\footnotesize 6.2}} & 68.0 \down{\textit{\footnotesize 3.4}} & 29.6 \up{\textit{\footnotesize 0.9}} \\
        Spatial Misalignment & 70.4 \down{\textit{\footnotesize 2.6}} & 73.6 \down{\textit{\footnotesize 1.4}} & 27.1 \up{\textit{\footnotesize 0.2}} & 61.2 \down{\textit{\footnotesize 7.3}} & 67.3 \down{\textit{\footnotesize 4.1}} & 29.8 \up{\textit{\footnotesize 1.1}} \\
        \bottomrule
        \end{tabular}}
    \label{tab:nusc_class}
\caption{Robustness to image depth estimation noise on nuScenes validation set.}
\label{tab:noise}
\end{table*}

\begin{table}[t]
    \centering
     \setlength{\tabcolsep}{1mm}
    {
    \begin{tabular}{lccc}
        \toprule
        Methods & Modality & NDS & mAP\\
        \midrule
        Baseline & LC & 73.3 & 70.6 \\
        \midrule
        + SVDB (w/o image) & LC & 73.6 & 71.3 \\
        + SVDB (zero init) & LC & 72.7 & 70.4\\
        + SVDB (ours) & LC & 74.0 & 71.8\\
        \midrule
        + SBDB (w/o image) & LC & 74.2 & 71.7\\
        + SBDB (fusion) & LC & 74.4 & 71.8 \\
        + SBDB (ours) & LC & 74.8 & 72.5\\
        \bottomrule
    \end{tabular}}
\caption{Ablation studies for our proposed components on the nuScenes \textbf{validation} set. `w/o image’ denotes the removal of image priors in the current block. `fusion’ means using DCN on multimodal features in SBDB.}
\label{tab:alter}
\end{table}

\subsection{Ablation Studies}
To better investigate the effectiveness of BEVDilation, we conduct a set of ablation studies by using the nuScenes validation set in this subsection. All the models are trained end-to-end for 10 epochs.

\noindent \textbf{Effectiveness of each component.} 
To more clearly demonstrate the effectiveness of the SVDB and SBDB in BEVDilation, we conduct experiments by adding each of them to a multi-modal baseline. 
The baseline takes VoxelNet~\cite{Voxelnet} and SECOND~\cite{Second} dense backbone as the LiDAR backbone and ResNet50~\cite{ResNet} as the image backbone. 
The detection head follows the design in DAL~\cite{huang2024DAL}.
As evidenced in Table~\ref{tab:eachcomp}, both SVDB (+1.2 mAP) and SBDB (+1.9 mAP) can significantly improve the accuracy over the baseline, which validates the feasibility of our LiDAR-centric strategy. 
Combining SVDB and SBDB further improves performance, suggesting that densifying sparse BEV foreground enhances feature diffusion, leading to better completion of LiDAR occluded regions.

\noindent \textbf{Robustness of our LiDAR-centric paradigm.} To validate the robustness and effectiveness of our LiDAR-centric paradigm against indiscriminate fusion, we conducted experiments between our BEVDilation and BEVFusion~\cite{liu2023bevfusion-mit}. We introduce three types of degradation during inference: depth random noise, depth one-hot noise~\cite{li2023bevdepth}, and spatial misalignment~\cite{dong2023benchmarking}. As shown in Table~\ref{tab:noise}, BEVDilation exhibits lower performance drops than BEVFusion under all degradations, demonstrating its superior robustness.
Furthermore, under noisy conditions, BEVDilation consistently achieves higher mATE than standard BEVFusion, indicating that prioritizing LiDAR information helps maintain accurate geometric perception.

\noindent \textbf{Effectiveness of multimodal features.}
To validate the efficacy of image semantic information in our LiDAR-centric framework, we perform experiments on some potential alternatives to SVDB and SBDB.
As shown in Table~\ref{tab:alter}, removing image guidance for SVDB foreground mask prediction leads to substantial performance drops, -0.5 mAP and -0.4 NDS, demonstrating the critical role of images in recovering LiDAR-occluded regions. 
Furthermore, replacing our learnable embedding for padded empty locations with a fixed zero value~\cite{liu2025lion,swformer} causes a significant performance drop of 1.2 mAP. This indicates that handcrafted features fail to adapt to scene-specific contexts and result in suboptimal feature representations.
For SBDB, removing image-conditioned deformations in DCN leads to a significant decline in performance, which highlights the necessity of semantic and context guidance from images.
We also compare our LiDAR-centric fusion within SBDB against a direct fusion baseline that uses standard deformable convolutions on concatenated multi-modal BEV features.
The decreased performance confirms that our approach can effectively reduce the impact of noise from image depth estimation.

\noindent \textbf{Visualization of BEVDilation.}
To validate the effectiveness of our semantic-guided BEV dilation, we visualize its sampling locations across different stages. 
We project the multi-modal conditioned sampling locations from all groups~\cite{wang2023DCNv3} onto a unified BEV map and filter out those locations with modulation scalars below 0.01.
As shown in Fig~\ref{fig:deformation}, compared to rigid grid patterns in standard convolutions, SBDB dynamically adapts its sampling locations to geometric variations.
For example, as shown in stage 3, SBDB can directly extract features at object boundaries, reducing the need for deep CNN stacks that struggle with off-grid points.
This clearly demonstrates SBDB’s ability to handle the irregular distributions in point clouds.
Furthermore, the progressively expanding receptive fields in the deeper stages highlight SBDB's capability to aggregate long-range contextual information, which is important to address the Center Feature Missing issue.
We further include visualizations of the predicted foreground occupancy from SVDB in the \textbf{supplementary material}.

\section{Conclusion}
\label{sec:conclusion}

In this paper, we proposed BEVDilation, a LiDAR-centric backbone for multi-modal 3D object detection. We first analyzed the degradation from current naive fusion. Thus, we proposed a LiDAR-centric paradigm to reduce the influence of image depth estimation noise in fusion.
In specific, the SVDB block is used to densify foreground voxel features and fill in the LiDAR-occluded regions with image priors.
Then, the SBDB equipped with multi-modal conditioned deformations diffuses sparse voxel features into dense BEV features with geometry-aware adaptability and enhanced semantic context.
Experiments demonstrated that BEVDilation achieved state-of-the-art results on the nuScenes dataset.
BEVDilation presented a new LiDAR-centric solution for multi-modal 3D detection.




\begin{figure*}[t]
\centering
{\LARGE \textbf{Supplementary Material}}
\end{figure*}

\begin{figure*}[t]
     \centering
    \includegraphics[width=0.8\textwidth]{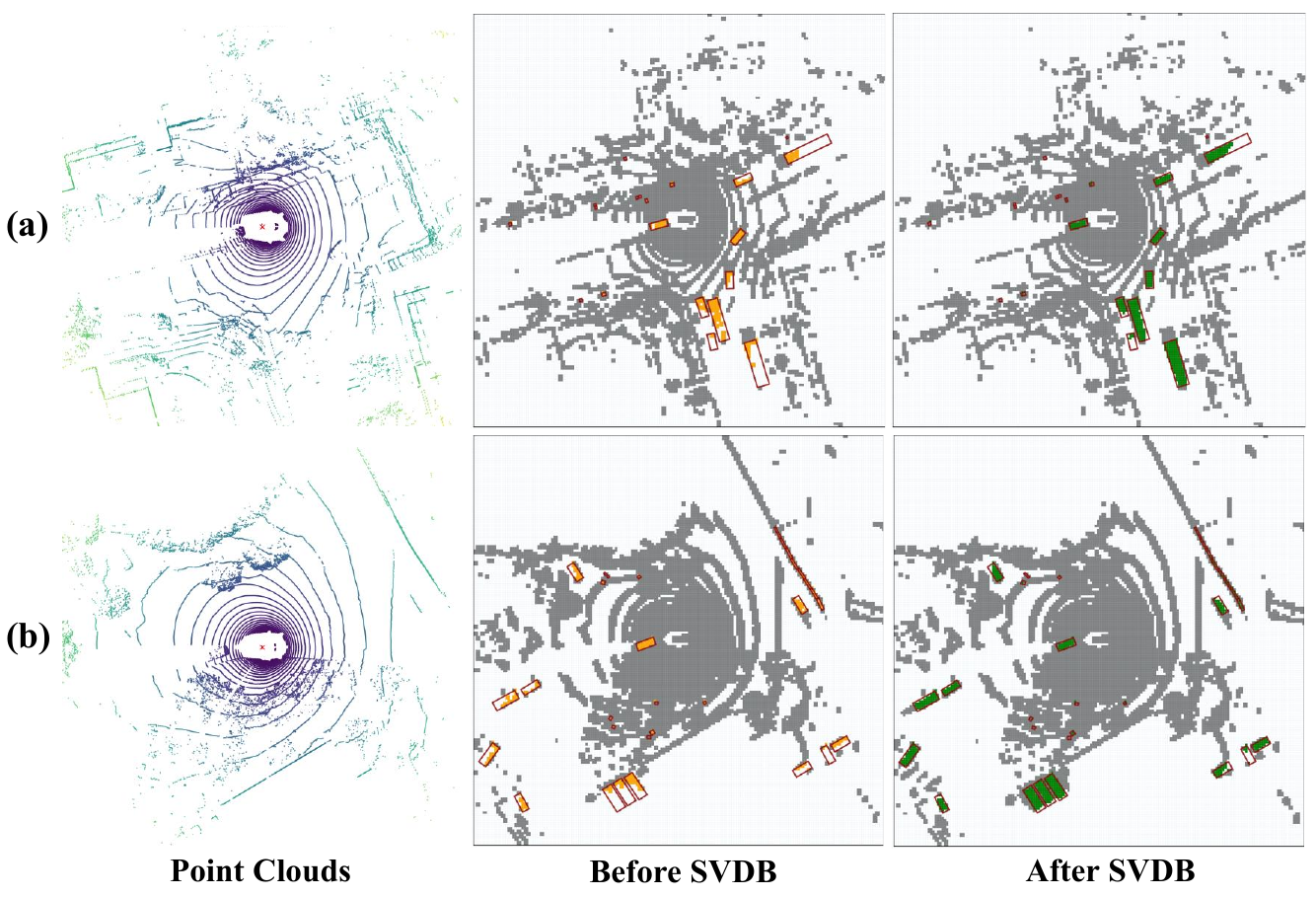}
     \caption{Visualization of the dilated and original occupancy. BEV grids within the ground truth bounding boxes are highlighted: orange indicates the original occupancy, while green denotes the dilated occupancy.
     }
    \label{fig:main}
\end{figure*}

\begin{table*}[!ht]
    \centering
    \setlength{\tabcolsep}{1.4mm}{}
    \scalebox{0.75}{
        \begin{tabular}{l|cc|ccccc|cccccccccc}
        \toprule
        Method & NDS & mAP & mATE$\downarrow$ & mASE$\downarrow$ & mAOE$\downarrow$ & mAVE$\downarrow$ & mAEE$\downarrow$ & Car & Truck &  C.V. & Bus & T.L. & B.R. & M.T. & Bike & Ped. & T.C.\\ 
        \midrule
        TransFusion~\cite{transfusion} & 71.7 & 68.9 & 25.9 & 24.3 & 32.9 & 28.8 & 12.7 & 87.1 & 60.0 & 33.1 & 68.3 & 60.8 & 78.1 & 73.6 & 52.9 & 88.4 & 86.7 \\
        BEVFusion~\cite{liang2022bevfusion-pku} & 71.8 & 69.2 & 25.0 & 24.0 & 35.9 & 25.4 & 13.2 & 88.1 & 60.9 & 34.4 & 69.3 & 62.1 & 78.2 & 72.2 & 52.2 & 89.2 & 85.5\\
        BEVFusion~\cite{liu2023bevfusion-mit} & 72.9 & 70.2 & 26.1 & 23.9 & 32.9 & 26.0 & 13.4 & 88.6 & 60.1 & 39.3 & 69.8 & 63.8 & 80.0 & 74.1 & 51.0 & 89.2 & 86.5 \\
        DeepInteraction~\cite{yang2022deepinteraction} & 73.4 & 70.8 & 25.7 & 24.0 & 32.5 & 24.5 & 12.8 & 87.9 & 60.2 & 37.5 & 70.8 & 63.8 & 80.4 & 75.4 & 54.5 & \textbf{91.7} & 87.2\\
        ObjectFusion~\cite{cai2023objectfusion} & 73.3 & 71.0 & - & - & - & - & - & 89.4 & 59.0 & \textbf{40.5} & 71.8 & 63.1 & 80.0 & 78.1 & 53.2 & 90.7 & 87.7\\
        MSMDFusion~\cite{jiao2023msmdfusion} & 74.0 & 71.5 & 25.5 & 23.8 & 31.0 & 24.4 & 13.2 & 88.4 & 61.0 & 35.2 & 71.4 & 64.2 & \textbf{80.7} & 76.9 & 58.3 & 90.6 & 88.1 \\
        GraphBEV~\cite{song2024graphbev} & 73.6 & 71.7 & - & - & - & - & - & 89.2 & 60.0 & 40.8 & 72.1 & 64.5 & 80.1 & 76.8 & 53.3 & 90.9 & \textbf{88.9}\\
        FocalFormer3D~\cite{chen2023focalformer3d} & 73.9 & 71.6 & - & - & - & - & - & 88.5 & 61.4 & 35.9 & 71.7 & 66.4 & 79.3 & 80.3 & 57.1 & 89.7 & 85.3 \\
        CMT~\cite{yan2023CMT} & 74.1 & 72.0 & 27.9 & 23.5 & 30.8 & 25.9 & 11.2 & 88.0 & 63.3 & 37.3 & {75.4} & {65.4} & 78.2 & 79.1 & \textbf{60.6} & 87.9 & 84.7 \\
        SparseFusion~\cite{xie2023sparsefusion} & 73.8 & 72.0 & 25.8 & 24.3 & 32.9 & 26.5 & 13.1 & 88.0 & 60.2 & 38.7 & 72.0 & 64.9 & 79.2 & 78.5 & 59.8 & 90.0 & 87.9 \\
        DAL~\cite{huang2024DAL} & 74.8 & 72.0 & 25.3 & 23.9 & 33.4 & 17.4 & 12.0 & 89.1 & 60.2 & 34.6 & 73.3 & 65.8  & 80.6 & \textbf{81.7} & 58.5 & 89.6 & 86.6  \\
        \midrule
        \textbf{BEVDilation} (ours) & \textbf{75.4} & \textbf{73.1} & \textbf{24.8} & \textbf{23.4} & 33.8 & 17.8 & 11.7 & \textbf{90.4} & \textbf{63.5} &  39.1 & \textbf{75.8} & \textbf{69.2} &  {80.6} & 77.7 & 57.1 & 90.5 & {87.3} \\
        \bottomrule
        \end{tabular}}
    \caption{Comparison with the state-of-the-art detectors on the nuScenes dataset \textbf{test} split. `C.V.', `T.L.', `B.R.', `M.T.', `Ped', and `T.C' denote construction vehicle, trailer, barrier, motor, pedestrian, and traffic cone, respectively.}
    \label{tab:nusc_class}
\end{table*}

\subsection{Additional Experiments}
As illustrated in Table~\ref{tab:nusc_class}, we report the detailed performance analysis, including metrics such as mATE and per-class mAP on the nuScenes dataset.
BEVDilation obtains the best results in most categories, with particularly notable improvements in classes that are sensitive to the feature diffusion (\textit{e.g.,} bus, truck, and trailer). 
Moreover, for categories with relatively small scale in the scene (\textit{e.g.,} pedestrian, barrier, and traffic cone), BEVDilation still achieves competitive performance.
It demonstrates that BEVDilation exhibits a certain degree of robustness to noise in foreground mask prediction.
Furthermore, compared to other methods, BEVDilation achieves higher mATE and mASE, which indicates that using the image as implicit guidance effectively reduces the reliance on explicit depth estimation and improves the algorithm's geometric perception accuracy.
This validates that BEVDilation successfully diffuses sparse voxel features into semantically rich and geometrically accurate BEV representations. 

To validate the generalizability of BEVDilation, we extended our experiments to the Waymo open dataset~\cite{Waymo}. Waymo open dataset contains 230k annotated samples, partitioned into 160k for training, 40k for validation, and 30k for testing. Each frame covers a large perception range $(150m \times 150m)$. The detection results are evaluated using the mean average precision (mAP) and its weighted variant, the mean average precision with heading accuracy (mAPH). These metrics are further categorized into Level 1 for objects detected by over five points, and Level 2 for those detected with at least one point.

  \begin{table}[t]
     \centering
     \resizebox{\linewidth}{!}{
      \setlength{\tabcolsep}{1mm} 
     \begin{tabular}{l|c|cccc}
        \toprule
        Methods & Modality & Overall & Vehicle & Pedestrian & Cyclist\\
        \midrule
        SECOND & L & 57.2 & 63.3 & 51.3 & 57.1 \\
        PointPillar & L & 57.8 & 63.1 & 50.3 & 59.9 \\
        TransFusion-L & L & 64.9 & 65.1 & 63.7 & 65.9\\
        CenterPoint & L & 67.6 & 68.4 & 65.8 & 68.5\\
        \midrule
        PointAugmenting & LC & 66.7 & 62.2 & 64.6 & \textbf{73.3}\\
        TransFusion & LC & 65.5 & 65.1 & 64.0 & 67.4\\
        DeepFusion & LC & 67.0 & - & - & -\\
        BEVDilation & LC & \textbf{70.1} & \textbf{69.6} & \textbf{68.8} & 71.8\\
        \bottomrule
    \end{tabular}
    }
    \caption{Results on Waymo \textbf{validation} set measured by LEVEL 2 mAPH. All the methods are under the single-frame setting for fair comparison.}
    \label{tab:waymo}
  \end{table}

For Waymo, BEVDilation is implemented based on the open-source framework OpenPCDet~\cite{openpcdet}. For the camera branch, we use Swin-Tiny~\cite{swin_transformer} as the image backbone and initialize it with weights pre-trained on nuImage~\cite{Nuscenes}. The input image resolution is $704 \times 256$. For the LiDAR branch, we follow CenterPoint~\cite{Centerpoint} and use the voxel size $(0.1m, 0.1m, 0.15m)$ with the point cloud range defined as $[-75.2m, -75.2m]$ along the XY-axes, and $[-5m, 3m]$ for the Z-axis. We adopt the sparse 3D backbone from CenterPoint~\cite{Centerpoint}, consistent with that in previous works~\cite{li2024FSF,li2022deepfusion}. For the 2D dense CNN, we also stack eight SBDB blocks distributed over four stages. The $\tau$ in SVDB is set to 0.4. The initial BEV feature map is of size $188\times 188$.
Following previous multi-modal training schemes~\cite{li2024FSF,li2023logonet,transfusion} in Waymo, the 3D sparse backbone from CenterPoint is first pretrained on the training set. Then, in the second stage, the whole BEVDilation is trained for 12 epochs. We optimize the model using AdamW optimizer with weight decay 0.01, one-cycle learning rate policy, max learning rate 0.0003, and batch size of 24. All the stages are trained on 8 RTX A6000 GPUs.
As shown in the Table~\ref{tab:waymo}, following previous work~\cite{li2024FSF,transfusion}, we report the performance of our model across all three classes on the Waymo validation set. BEVDilation also achieves comparable results to the state-of-the-art methods. 

\subsection{Additional Visualization}
To validate the effectiveness of our Sparse Voxel Dilation Blocks (SVDB), we visualize the dilated and original occupancy. 
As shown in Fig.~\ref{fig:main}, we present visualizations of the original point clouds, their corresponding BEV occupancy maps, and the BEV occupancy maps after SVDB padding.
To better illustrate the dilation effect, the grids within the object bounding boxes are highlighted.
Figs.~\ref{fig:main} (a) and (b) clearly demonstrate that SVDB is capable of diffusing features to missing object centers, thereby densifying foreground features and improving feature representation in highly sparse point clouds. 
This finding accounts for the excellent performance of our method on large-scale objects (\textit{e.g.,} bus, truck, and trailer).
Furthermore, while some objects may not be fully reconstructed with foreground points, the successful localization of their centers still significantly benefits the optimization of BEVDilation.
However, as shown in Fig.~\ref{fig:main} (a), some fully occluded objects remain undetectable, which motivates the subsequent SBDB module for further feature diffusion.
In conclusion, by leveraging complementary multi-modal features, the Sparse Voxel Dilation Block (SVDB) is capable of identifying foreground regions in BEV space and populating sparse foreground areas in LiDAR data with learnable voxels.

\bibliography{aaai2026}

@inproceedings{Voxelnet,
  title={Voxelnet: End-to-end learning for point cloud based 3d object detection},
  author={Zhou, Yin and Tuzel, Oncel},
  booktitle={Proceedings of the IEEE conference on computer vision and pattern recognition},
  pages={4490--4499},
  year={2018}
}

@inproceedings{Centerpoint,
  title={Center-based 3d object detection and tracking},
  author={Yin, Tianwei and Zhou, Xingyi and Krahenbuhl, Philipp},
  booktitle={Proceedings of the IEEE/CVF conference on computer vision and pattern recognition},
  pages={11784--11793},
  year={2021}
}

@inproceedings{Pointnet,
  title={Pointnet: Deep learning on point sets for 3d classification and segmentation},
  author={Qi, Charles R and Su, Hao and Mo, Kaichun and Guibas, Leonidas J},
  booktitle={Proceedings of the IEEE conference on computer vision and pattern recognition},
  pages={652--660},
  year={2017}
}

@inproceedings{Pointrcnn,
  title={Pointrcnn: 3d object proposal generation and detection from point cloud},
  author={Shi, Shaoshuai and Wang, Xiaogang and Li, Hongsheng},
  booktitle={Proceedings of the IEEE/CVF conference on computer vision and pattern recognition},
  pages={770--779},
  year={2019}
}

@inproceedings{Votenet,
  title={Deep hough voting for 3d object detection in point clouds},
  author={Qi, Charles R and Litany, Or and He, Kaiming and Guibas, Leonidas J},
  booktitle={proceedings of the IEEE/CVF International Conference on Computer Vision},
  pages={9277--9286},
  year={2019}
}

@inproceedings{Nuscenes,
  title={nuscenes: A multimodal dataset for autonomous driving},
  author={Caesar, Holger and Bankiti, Varun and Lang, Alex H and Vora, Sourabh and Liong, Venice Erin and Xu, Qiang and Krishnan, Anush and Pan, Yu and Baldan, Giancarlo and Beijbom, Oscar},
  booktitle={Proceedings of the IEEE/CVF conference on computer vision and pattern recognition},
  pages={11621--11631},
  year={2020}
}

@inproceedings{Waymo,
  title={Scalability in perception for autonomous driving: Waymo open dataset},
  author={Sun, Pei and Kretzschmar, Henrik and Dotiwalla, Xerxes and Chouard, Aurelien and Patnaik, Vijaysai and Tsui, Paul and Guo, James and Zhou, Yin and Chai, Yuning and Caine, Benjamin and others},
  booktitle={Proceedings of the IEEE/CVF conference on computer vision and pattern recognition},
  pages={2446--2454},
  year={2020}
}

@inproceedings{ResNet,
  title={Deep residual learning for image recognition},
  author={He, Kaiming and Zhang, Xiangyu and Ren, Shaoqing and Sun, Jian},
  booktitle={Proceedings of the IEEE conference on computer vision and pattern recognition},
  pages={770--778},
  year={2016}
}

@misc{mmdet3d,
    title={{MMDetection3D: OpenMMLab} next-generation platform for general {3D} object detection},
    author={MMDetection3D Contributors},
    howpublished = {\url{https://github.com/open-mmlab/mmdetection3d}},
    year={2020}
}

@article{Second,
  title={Second: Sparsely embedded convolutional detection},
  author={Yan, Yan and Mao, Yuxing and Li, Bo},
  journal={Sensors},
  volume={18},
  number={10},
  pages={3337},
  year={2018},
  publisher={MDPI}
}

@inproceedings{VoxelTransformer,
  title={Voxel transformer for 3d object detection},
  author={Mao, Jiageng and Xue, Yujing and Niu, Minzhe and Bai, Haoyue and Feng, Jiashi and Liang, Xiaodan and Xu, Hang and Xu, Chunjing},
  booktitle={Proceedings of the IEEE/CVF International Conference on Computer Vision},
  pages={3164--3173},
  year={2021}
}

@article{hilbertcurve,
  title={{\"U}ber die stetige Abbildung einer Linie auf ein Fl{\"a}chenst{\"u}ck},
  author={Hilbert, David and Hilbert, David},
  journal={Dritter Band: Analysis{\textperiodcentered} Grundlagen der Mathematik{\textperiodcentered} Physik Verschiedenes: Nebst Einer Lebensgeschichte},
  pages={1--2},
  year={1935},
  publisher={Springer}
}

@article{Mamba,
  title={Mamba: Linear-time sequence modeling with selective state spaces},
  author={Gu, Albert and Dao, Tri},
  journal={arXiv preprint arXiv:2312.00752},
  year={2023}
}

@inproceedings{transfusion,
  title={Transfusion: Robust lidar-camera fusion for 3d object detection with transformers},
  author={Bai, Xuyang and Hu, Zeyu and Zhu, Xinge and Huang, Qingqiu and Chen, Yilun and Fu, Hongbo and Tai, Chiew-Lan},
  booktitle={Proceedings of the IEEE/CVF conference on computer vision and pattern recognition},
  pages={1090--1099},
  year={2022}
}

@article{FSD,
  title={Fully sparse 3d object detection},
  author={Fan, Lue and Wang, Feng and Wang, Naiyan and Zhang, Zhao-Xiang},
  journal={Advances in Neural Information Processing Systems},
  volume={35},
  pages={351--363},
  year={2022}
}

@inproceedings{swformer,
  title={Swformer: Sparse window transformer for 3d object detection in point clouds},
  author={Sun, Pei and Tan, Mingxing and Wang, Weiyue and Liu, Chenxi and Xia, Fei and Leng, Zhaoqi and Anguelov, Dragomir},
  booktitle={European Conference on Computer Vision},
  pages={426--442},
  year={2022},
  organization={Springer}
}

@article{hednet,
  title={Hednet: A hierarchical encoder-decoder network for 3d object detection in point clouds},
  author={Zhang, Gang and Junnan, Chen and Gao, Guohuan and Li, Jianmin and Hu, Xiaolin},
  journal={Advances in Neural Information Processing Systems},
  volume={36},
  pages={53076--53089},
  year={2023}
}

@inproceedings{focalformer3d,
  title={Focalformer3d: focusing on hard instance for 3d object detection},
  author={Chen, Yilun and Yu, Zhiding and Chen, Yukang and Lan, Shiyi and Anandkumar, Anima and Jia, Jiaya and Alvarez, Jose M},
  booktitle={Proceedings of the IEEE/CVF International Conference on Computer Vision},
  pages={8394--8405},
  year={2023}
}

@article{sparsefusion,
  title={SparseFusion: Efficient Sparse Multi-Modal Fusion Framework for Long-Range 3D Perception},
  author={Li, Yiheng and Li, Hongyang and Huang, Zehao and Chang, Hong and Wang, Naiyan},
  journal={arXiv preprint arXiv:2403.10036},
  year={2024}
}

@inproceedings{swin_transformer,
  title={Swin transformer: Hierarchical vision transformer using shifted windows},
  author={Liu, Ze and Lin, Yutong and Cao, Yue and Hu, Han and Wei, Yixuan and Zhang, Zheng and Lin, Stephen and Guo, Baining},
  booktitle={Proceedings of the IEEE/CVF international conference on computer vision},
  pages={10012--10022},
  year={2021}
}

@article{UVTR,
  title={Unifying voxel-based representation with transformer for 3d object detection},
  author={Li, Yanwei and Chen, Yilun and Qi, Xiaojuan and Li, Zeming and Sun, Jian and Jia, Jiaya},
  journal={Advances in Neural Information Processing Systems},
  volume={35},
  pages={18442--18455},
  year={2022}
}

@misc{openpcdet,
    title={OpenPCDet: An Open-source Toolbox for 3D Object Detection from Point Clouds},
    author={OpenPCDet Development Team},
    howpublished = {\url{https://github.com/open-mmlab/OpenPCDet}},
    year={2020}
}

@article{e2ead_survey,
  title={End-to-end autonomous driving: Challenges and frontiers},
  author={Chen, Li and Wu, Penghao and Chitta, Kashyap and Jaeger, Bernhard and Geiger, Andreas and Li, Hongyang},
  journal={arXiv preprint arXiv:2306.16927},
  year={2023}
}

@inproceedings{VR_3Ddetection,
  title={Multiple 3d object tracking for augmented reality},
  author={Park, Youngmin and Lepetit, Vincent and Woo, Woontack},
  booktitle={2008 7th IEEE/ACM International Symposium on Mixed and Augmented Reality},
  pages={117--120},
  year={2008},
  organization={IEEE}
}

@article{scatterformer,
  title={ScatterFormer: Efficient Voxel Transformer with Scattered Linear Attention},
  author={He, Chenhang and Li, Ruihuang and Zhang, Guowen and Zhang, Lei},
  journal={arXiv preprint arXiv:2401.00912},
  year={2024}
}

@article{GGA,
  title={General Geometry-aware Weakly Supervised 3D Object Detection},
  author={Zhang, Guowen and Fan, Junsong and Chen, Liyi and Zhang, Zhaoxiang and Lei, Zhen and Zhang, Lei},
  journal={arXiv preprint arXiv:2407.13748},
  year={2024}
}

@inproceedings{wang2021pointaugmenting,
  title={Pointaugmenting: Cross-modal augmentation for 3d object detection},
  author={Wang, Chunwei and Ma, Chao and Zhu, Ming and Yang, Xiaokang},
  booktitle={Proceedings of the IEEE/CVF conference on computer vision and pattern recognition},
  pages={11794--11803},
  year={2021}
}

@inproceedings{chen2023futr3d,
  title={Futr3d: A unified sensor fusion framework for 3d detection},
  author={Chen, Xuanyao and Zhang, Tianyuan and Wang, Yue and Wang, Yilun and Zhao, Hang},
  booktitle={proceedings of the IEEE/CVF conference on computer vision and pattern recognition},
  pages={172--181},
  year={2023}
}

@inproceedings{chen2022autoalignv2,
  title={Deformable feature aggregation for dynamic multi-modal 3D object detection},
  author={Chen, Zehui and Li, Zhenyu and Zhang, Shiquan and Fang, Liangji and Jiang, Qinhong and Zhao, Feng},
  booktitle={European conference on computer vision},
  pages={628--644},
  year={2022},
  organization={Springer}
}

@article{liang2022bevfusion-pku,
  title={Bevfusion: A simple and robust lidar-camera fusion framework},
  author={Liang, Tingting and Xie, Hongwei and Yu, Kaicheng and Xia, Zhongyu and Lin, Zhiwei and Wang, Yongtao and Tang, Tao and Wang, Bing and Tang, Zhi},
  journal={Advances in Neural Information Processing Systems},
  volume={35},
  pages={10421--10434},
  year={2022}
}

@inproceedings{liu2023bevfusion-mit,
  title={Bevfusion: Multi-task multi-sensor fusion with unified bird's-eye view representation},
  author={Liu, Zhijian and Tang, Haotian and Amini, Alexander and Yang, Xinyu and Mao, Huizi and Rus, Daniela L and Han, Song},
  booktitle={2023 IEEE international conference on robotics and automation (ICRA)},
  pages={2774--2781},
  year={2023},
  organization={IEEE}
}

@article{yang2022deepinteraction,
  title={Deepinteraction: 3d object detection via modality interaction},
  author={Yang, Zeyu and Chen, Jiaqi and Miao, Zhenwei and Li, Wei and Zhu, Xiatian and Zhang, Li},
  journal={Advances in Neural Information Processing Systems},
  volume={35},
  pages={1992--2005},
  year={2022}
}

@inproceedings{yan2023CMT,
  title={Cross modal transformer: Towards fast and robust 3d object detection},
  author={Yan, Junjie and Liu, Yingfei and Sun, Jianjian and Jia, Fan and Li, Shuailin and Wang, Tiancai and Zhang, Xiangyu},
  booktitle={Proceedings of the IEEE/CVF international conference on computer vision},
  pages={18268--18278},
  year={2023}
}

@inproceedings{huang2024DAL,
  title={Detecting as labeling: Rethinking lidar-camera fusion in 3d object detection},
  author={Huang, Junjie and Ye, Yun and Liang, Zhujin and Shan, Yi and Du, Dalong},
  booktitle={European Conference on Computer Vision},
  pages={439--455},
  year={2024},
  organization={Springer}
}

@inproceedings{wang2023unitr,
  title={Unitr: A unified and efficient multi-modal transformer for bird's-eye-view representation},
  author={Wang, Haiyang and Tang, Hao and Shi, Shaoshuai and Li, Aoxue and Li, Zhenguo and Schiele, Bernt and Wang, Liwei},
  booktitle={Proceedings of the IEEE/CVF international conference on computer vision},
  pages={6792--6802},
  year={2023}
}

@inproceedings{chen2023focalformer3d,
  title={Focalformer3d: focusing on hard instance for 3d object detection},
  author={Chen, Yilun and Yu, Zhiding and Chen, Yukang and Lan, Shiyi and Anandkumar, Anima and Jia, Jiaya and Alvarez, Jose M},
  booktitle={Proceedings of the IEEE/CVF International Conference on Computer Vision},
  pages={8394--8405},
  year={2023}
}

@inproceedings{yin2024isfusion,
  title={Is-fusion: Instance-scene collaborative fusion for multimodal 3d object detection},
  author={Yin, Junbo and Shen, Jianbing and Chen, Runnan and Li, Wei and Yang, Ruigang and Frossard, Pascal and Wang, Wenguan},
  booktitle={Proceedings of the IEEE/CVF conference on computer vision and pattern recognition},
  pages={14905--14915},
  year={2024}
}

@inproceedings{xie2023sparsefusion,
  title={Sparsefusion: Fusing multi-modal sparse representations for multi-sensor 3d object detection},
  author={Xie, Yichen and Xu, Chenfeng and Rakotosaona, Marie-Julie and Rim, Patrick and Tombari, Federico and Keutzer, Kurt and Tomizuka, Masayoshi and Zhan, Wei},
  booktitle={Proceedings of the IEEE/CVF International Conference on Computer Vision},
  pages={17591--17602},
  year={2023}
}

@inproceedings{cai2023objectfusion,
  title={Objectfusion: Multi-modal 3d object detection with object-centric fusion},
  author={Cai, Qi and Pan, Yingwei and Yao, Ting and Ngo, Chong-Wah and Mei, Tao},
  booktitle={Proceedings of the IEEE/CVF International Conference on Computer Vision},
  pages={18067--18076},
  year={2023}
}

@inproceedings{jiao2023msmdfusion,
  title={Msmdfusion: Fusing lidar and camera at multiple scales with multi-depth seeds for 3d object detection},
  author={Jiao, Yang and Jie, Zequn and Chen, Shaoxiang and Chen, Jingjing and Ma, Lin and Jiang, Yu-Gang},
  booktitle={Proceedings of the IEEE/CVF conference on computer vision and pattern recognition},
  pages={21643--21652},
  year={2023}
}

@inproceedings{li2023bevdepth,
  title={Bevdepth: Acquisition of reliable depth for multi-view 3d object detection},
  author={Li, Yinhao and Ge, Zheng and Yu, Guanyi and Yang, Jinrong and Wang, Zengran and Shi, Yukang and Sun, Jianjian and Li, Zeming},
  booktitle={Proceedings of the AAAI conference on artificial intelligence},
  volume={37},
  number={2},
  pages={1477--1485},
  year={2023}
}

@article{li2024bevformer,
  title={Bevformer: learning bird's-eye-view representation from lidar-camera via spatiotemporal transformers},
  author={Li, Zhiqi and Wang, Wenhai and Li, Hongyang and Xie, Enze and Sima, Chonghao and Lu, Tong and Yu, Qiao and Dai, Jifeng},
  journal={IEEE Transactions on Pattern Analysis and Machine Intelligence},
  year={2024},
  publisher={IEEE}
}

@inproceedings{liu2023petrv2,
  title={Petrv2: A unified framework for 3d perception from multi-camera images},
  author={Liu, Yingfei and Yan, Junjie and Jia, Fan and Li, Shuailin and Gao, Aqi and Wang, Tiancai and Zhang, Xiangyu},
  booktitle={Proceedings of the IEEE/CVF International Conference on Computer Vision},
  pages={3262--3272},
  year={2023}
}

@inproceedings{liu2022petr,
  title={Petr: Position embedding transformation for multi-view 3d object detection},
  author={Liu, Yingfei and Wang, Tiancai and Zhang, Xiangyu and Sun, Jian},
  booktitle={European conference on computer vision},
  pages={531--548},
  year={2022},
  organization={Springer}
}

@inproceedings{pointpainting,
  title={Pointpainting: Sequential fusion for 3d object detection},
  author={Vora, Sourabh and Lang, Alex H and Helou, Bassam and Beijbom, Oscar},
  booktitle={Proceedings of the IEEE/CVF conference on computer vision and pattern recognition},
  pages={4604--4612},
  year={2020}
}

@article{yin2021multimodal,
  title={Multimodal virtual point 3d detection},
  author={Yin, Tianwei and Zhou, Xingyi and Kr{\"a}henb{\"u}hl, Philipp},
  journal={Advances in Neural Information Processing Systems},
  volume={34},
  pages={16494--16507},
  year={2021}
}

@article{liu2025lion,
  title={Lion: Linear group rnn for 3d object detection in point clouds},
  author={Liu, Zhe and Hou, Jinghua and Wang, Xinyu and Ye, Xiaoqing and Wang, Jingdong and Zhao, Hengshuang and Bai, Xiang},
  journal={Advances in Neural Information Processing Systems},
  volume={37},
  pages={13601--13626},
  year={2025}
}

@article{zhang2025voxelmamba,
  title={Voxel mamba: Group-free state space models for point cloud based 3d object detection},
  author={Zhang, Guowen and Fan, Lue and He, Chenhang and Lei, Zhen and ZHANG, ZHAO-XIANG and Zhang, Lei},
  journal={Advances in Neural Information Processing Systems},
  volume={37},
  pages={81489--81509},
  year={2024}
}

@inproceedings{zhang2024safdnet,
  title={Safdnet: A simple and effective network for fully sparse 3d object detection},
  author={Zhang, Gang and Chen, Junnan and Gao, Guohuan and Li, Jianmin and Liu, Si and Hu, Xiaolin},
  booktitle={Proceedings of the IEEE/CVF Conference on Computer Vision and Pattern Recognition},
  pages={14477--14486},
  year={2024}
}

@inproceedings{zhu2019DCNv2,
  title={Deformable convnets v2: More deformable, better results},
  author={Zhu, Xizhou and Hu, Han and Lin, Stephen and Dai, Jifeng},
  booktitle={Proceedings of the IEEE/CVF conference on computer vision and pattern recognition},
  pages={9308--9316},
  year={2019}
}

@inproceedings{dai2017DCN,
  title={Deformable convolutional networks},
  author={Dai, Jifeng and Qi, Haozhi and Xiong, Yuwen and Li, Yi and Zhang, Guodong and Hu, Han and Wei, Yichen},
  booktitle={Proceedings of the IEEE international conference on computer vision},
  pages={764--773},
  year={2017}
}

@inproceedings{wang2023DCNv3,
  title={Internimage: Exploring large-scale vision foundation models with deformable convolutions},
  author={Wang, Wenhai and Dai, Jifeng and Chen, Zhe and Huang, Zhenhang and Li, Zhiqi and Zhu, Xizhou and Hu, Xiaowei and Lu, Tong and Lu, Lewei and Li, Hongsheng and others},
  booktitle={Proceedings of the IEEE/CVF conference on computer vision and pattern recognition},
  pages={14408--14419},
  year={2023}
}

@inproceedings{song2023graphalign,
  title={GraphAlign: Enhancing accurate feature alignment by graph matching for multi-modal 3D object detection},
  author={Song, Ziying and Wei, Haiyue and Bai, Lin and Yang, Lei and Jia, Caiyan},
  booktitle={Proceedings of the IEEE/CVF international conference on computer vision},
  pages={3358--3369},
  year={2023}
}

@article{zhu2019CBGS,
  title={Class-balanced grouping and sampling for point cloud 3d object detection},
  author={Zhu, Benjin and Jiang, Zhengkai and Zhou, Xiangxin and Li, Zeming and Yu, Gang},
  journal={arXiv preprint arXiv:1908.09492},
  year={2019}
}

@inproceedings{lin2017FPN,
  title={Feature pyramid networks for object detection},
  author={Lin, Tsung-Yi and Doll{\'a}r, Piotr and Girshick, Ross and He, Kaiming and Hariharan, Bharath and Belongie, Serge},
  booktitle={Proceedings of the IEEE conference on computer vision and pattern recognition},
  pages={2117--2125},
  year={2017}
}

@inproceedings{philion2020LSS,
  title={Lift, splat, shoot: Encoding images from arbitrary camera rigs by implicitly unprojecting to 3d},
  author={Philion, Jonah and Fidler, Sanja},
  booktitle={Computer Vision--ECCV 2020: 16th European Conference, Glasgow, UK, August 23--28, 2020, Proceedings, Part XIV 16},
  pages={194--210},
  year={2020},
  organization={Springer}
}

@inproceedings{song2024graphbev,
  title={Graphbev: Towards robust bev feature alignment for multi-modal 3d object detection},
  author={Song, Ziying and Yang, Lei and Xu, Shaoqing and Liu, Lin and Xu, Dongyang and Jia, Caiyan and Jia, Feiyang and Wang, Li},
  booktitle={European Conference on Computer Vision},
  pages={347--366},
  year={2024},
  organization={Springer}
}

@inproceedings{dong2023benchmarking,
  title={Benchmarking robustness of 3d object detection to common corruptions},
  author={Dong, Yinpeng and Kang, Caixin and Zhang, Jinlai and Zhu, Zijian and Wang, Yikai and Yang, Xiao and Su, Hang and Wei, Xingxing and Zhu, Jun},
  booktitle={Proceedings of the IEEE/CVF Conference on Computer Vision and Pattern Recognition},
  pages={1022--1032},
  year={2023}
}

@inproceedings{zhao2024simdistill,
  title={Simdistill: Simulated multi-modal distillation for bev 3d object detection},
  author={Zhao, Haimei and Zhang, Qiming and Zhao, Shanshan and Chen, Zhe and Zhang, Jing and Tao, Dacheng},
  booktitle={Proceedings of the AAAI conference on artificial intelligence},
  volume={38},
  number={7},
  pages={7460--7468},
  year={2024}
}

@inproceedings{li2022deepfusion,
  title={Deepfusion: Lidar-camera deep fusion for multi-modal 3d object detection},
  author={Li, Yingwei and Yu, Adams Wei and Meng, Tianjian and Caine, Ben and Ngiam, Jiquan and Peng, Daiyi and Shen, Junyang and Lu, Yifeng and Zhou, Denny and Le, Quoc V and others},
  booktitle={Proceedings of the IEEE/CVF conference on computer vision and pattern recognition},
  pages={17182--17191},
  year={2022}
}

@inproceedings{li2023logonet,
  title={Logonet: Towards accurate 3d object detection with local-to-global cross-modal fusion},
  author={Li, Xin and Ma, Tao and Hou, Yuenan and Shi, Botian and Yang, Yuchen and Liu, Youquan and Wu, Xingjiao and Chen, Qin and Li, Yikang and Qiao, Yu and others},
  booktitle={Proceedings of the IEEE/CVF conference on computer vision and pattern recognition},
  pages={17524--17534},
  year={2023}
}

@article{li2024FSF,
  title={Fully sparse fusion for 3d object detection},
  author={Li, Yingyan and Fan, Lue and Liu, Yang and Huang, Zehao and Chen, Yuntao and Wang, Naiyan and Zhang, Zhaoxiang},
  journal={IEEE Transactions on Pattern Analysis and Machine Intelligence},
  volume={46},
  number={11},
  pages={7217--7231},
  year={2024},
  publisher={IEEE}
}

@article{liyi2025vip3de,
  title={Fast Multi-view Consistent 3D Editing with Video Priors},
  author={Chen, Liyi and Li, Ruihuang and Zhang, Guowen and Wang, Pengfei and Zhang, Lei},
  journal={arXiv preprint arXiv:2511.23172},
  year={2025}
}

@article{man2025locateanything3d,
  title={LocateAnything3D: Vision-Language 3D Detection with Chain-of-Sight},
  author={Man, Yunze and Wang, Shihao and Zhang, Guowen and Bjorck, Johan and Li, Zhiqi and Gui, Liang-Yan and Fan, Jim and Kautz, Jan and Wang, Yu-Xiong and Yu, Zhiding},
  journal={arXiv preprint arXiv:2511.20648},
  year={2025}
}

\end{document}